ORIGINAL RESEARCH

# Obstacle-transformer: A trajectory prediction network based on surrounding trajectories

Wendong Zhang | Qingjie Chai | Quanqi Zhang | Chengwei Wu

Department of Control Science and Engineer, Harbin Institute of Technology, Harbin, China

**Correspondence**
Chengwei Wu, Department of Control Science and Engineer, Harbin Institute of Technology, Harbin, 150001, China.
Email: chengweiwu@hit.edu.cn

**Funding information**
National Natural Science Foundation of China, Grant/Award Number: 62033005

**Abstract**
Recurrent Neural Network, Long Short-Term Memory, and Transformer have made great progress in predicting the trajectories of moving objects. Although the trajectory element with the surrounding scene features has been merged to improve performance, there still exist some problems to be solved. One is that the time series processing models will increase the inference time with the increase of the number of prediction sequences. Another problem is that the features cannot be extracted from the scene's image and point cloud in some situations. Therefore, an Obstacle-Transformer is proposed to predict trajectory in a constant inference time. An 'obstacle' is designed by the surrounding trajectory rather than images or point clouds, making Obstacle-Transformer more applicable in a wider range of scenarios. Experiments are conducted on ETH and UCY datasets to verify the performance of our model.

## 1 | INTRODUCTION

With the recent development of autonomous driving and unmanned systems, trajectory prediction has been steadily gaining attention from researchers. Predicting trajectory becomes more and more crucial for the safety and subsequent decision-making for the autonomous system as well as other circumstances. Additionally, for the real-time physical system (such as autonomous vehicles), predicting trajectory in a highly accurate and real-time performance can provide a better reference for routine planning and kinematic control [1, 2].

In the past few decades, traditional methods of trajectory prediction have used mathematical methods to fit trajectory curves [3, 4]. With the development of deep-learning and natural language processing, time-series models have been proposed to process timing information. Currently, Recurrent Neural Network [5], Long Short-Term Memory (LSTM) [6], and Transformer [7] have been widely used in this field. However, these algorithms have some limitations. They generally make the inferences step-by-step. Each trajectory is modelled separately without interaction between the pedestrians and the scenes. Additionally, in some recent models, scene features are extracted from images and point clouds and integrated into the neural network. However, these methods are restricted to specific environmental feature inputs and are not suitable for situations in which scene images and point clouds do not exist.

In this paper, Obstacle-Transformer is proposed to solve the above problems. 'Obstacle-Transformer' is a neural network model that is utilised to predict pedestrians' trajectories. Transformer is the backbone of the network due to its capability of feature extraction and parallel computing. 'Obstacle' means that the model integrates surrounding obstacles' features by a method. To integrate features with the surrounding scenes, the 'obstacle' module based only on the recent past surrounding trajectories is proposed. The 'obstacle' is not limited by the real environment (e.g. images and point clouds), which makes the model have a better generality. Also, a Multi-Layer Perception (MLP) is designed in the model to predict the trajectory, which can output multiple time state values at the same time. Compared with other time series models (e.g. LSTM and raw Transformer [8]), the inference time can be optimised to a constant time and independent of the prediction length. The steps of training and prediction are shown in Figure 1. Experiments are conducted on pedestrian trajectory datasets to evaluate the performance of Obstacle-Transformer. The main contributions can be summarised as follows:

1. Obstacle-Transformer is proposed to predict trajectories. Compared with the existing results, for example, [8, 9], both the accuracy and real-time performance are improved.
2. The 'obstacle' is designed to integrate the surrounding scene features into the model. The use of 'obstacle' can get rid of the dependence on images or point clouds of the real environment.
3. The network architecture is designed to predict trajectory in a constant inference time, which ensures a real-time performance regardless of the predicted length.

## 2 | RELATED WORK

Trajectory prediction has been studied for decades. Algorithms in this field can be divided into two main trends. One trend is the model without the surrounding scene features integrated. In this trend, the trajectory is treated as a simple time series without interactive messages, so time-series models are used to process sequences. The other is to use a deep-learning model combined with the surrounding scene features to predict trajectories.

### 2.1 | Model without scene features

The studies of predicting trajectory without the surrounding scene features have been ongoing for decades, and the development of the technology has been improving.

Linear model [3], Gaussian regression model [10, 11], time-series analysis [12], and auto-regression model [13] were used in early work to predict the trajectories. These methods can perform well in simple scenes. However, once the scenes and the surrounding trajectories are complex, the fitness capability of these algorithms cannot be guaranteed.

With the development of deep learning, several attempts, such as LSTM [14] and Transformer [7, 8], have been made to predict trajectories. Yao et al. [15] proposed an end-to-end transformer network embedded with random deviation queries for pedestrian trajectory forecasting. These neural network models have a stronger fitting ability to fit non-linear trajectories. Meanwhile, several researchers have reported inspired optimizations of these models. Rae et al. [16] presented an attentive sequence model, Compressive Transformer, to compress memories for long-range sequence learning. Beltagy et al. [17] introduced Longformer with an attention mechanism that scales linearly with sequence length, making it easy to process documents with thousands of tokens or longer. Zhou et al. [18] proposed a ProbSparse Self-attention mechanism, which achieved $O(L \log L)$ in time complexity and memory usage.

### 2.2 | Model with scene features

Benefiting from the powerful flexibility and fitting capabilities of deep learning models, investigators have recently integrated the surrounding scene features into neural networks.

Some algorithms predict plausible trajectories taking into account a history of human motion paths. LealTaixé et al. [19] presented an approach to integrate the interaction by considering a social and grouping behaviour and using a global optimization scheme to solve the data association problem. Alahi et al. [20] proposed a new descriptor coined Social Affinity Maps to link broken or unobserved trajectories of individuals in the crowd. Fernando et al. [21] proposed a method that combines an attention model, which utilised both "soft attention" and "hard-wired" attention to map the trajectory information from the local neighbourhood to the future positions of the pedestrian of interest. Alahi et al. [9] proposed an LSTM model, which can learn general human movements and predict their future trajectories. Gupta et al. [22] predicted socially plausible futures by training adversarially against a recurrent discriminator and encouraged diverse predictions with a novel variety loss. Vemula et al. [23] proposed Social Attention, a novel trajectory prediction model that captures the relative importance of each person. Zhao et al. [24] proposed a novel spatial-temporal attention model to study spatial and temporal affinities jointly.

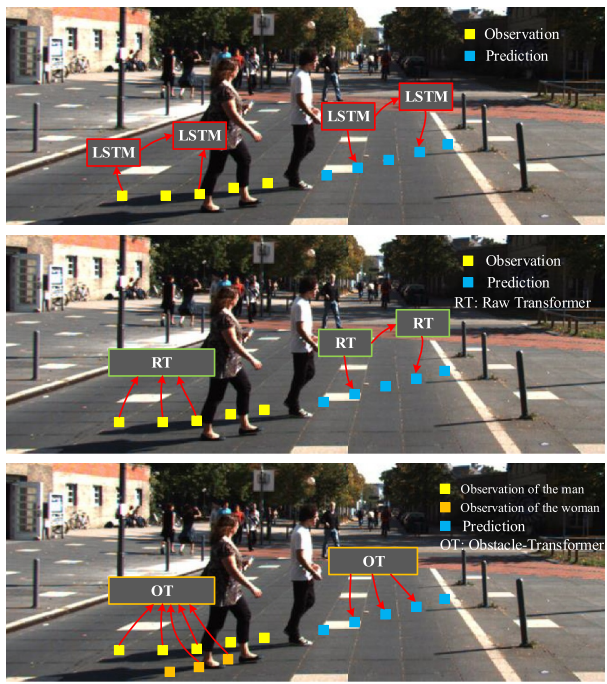

**FIGURE 1** A comparison of long short-term memory (LSTM), raw Transformer, and Obstacle-Transformer. The yellow dots are the observations and the blue dots are the predictions. LSTM sequentially processes the observations and sequentially predicts. The raw Transformer parallel processes the observations but sequentially predicts. The Obstacle-Transformer parallel processes the observations and makes a prediction

Meanwhile, some algorithms fuse multi-modal information and combine multiple scene features for trajectory prediction. Sadeghian et al. proposed an attentive Generative Adversarial Network (GAN) for predicting paths compliant to social and physical constraints (SoPhie) [25], which blended a social attention mechanism with physical attention and extracted the features of related images. Salzmann et al. [26] incorporated semantic maps and camera images for motion prediction. Wang et al. proposed a point-wise motion learning network (PointMotionNet) [27], which learns motion information from a sequence of large-scale 3D LiDAR point clouds. Peri et al. proposed a network that forecasts trajectories from LiDAR via future object detection [28].

## 3 | OBSTACLE TRANSFORMER MODEL

### 3.1 | Problem formulation

The pedestrians' trajectories are affected by their specific destination and the surroundings in the real situation. For the non-linear trajectories to be predicted, they are determined by several recent states, which are expressed as follows:

$$P[S_{t+1} \cdots S_{t+n} | S_t \cdots S_{t-m}] = P[S_{t+1} \cdots S_{t+n} | S_t \cdots S_{t-m-k}]$$

where $S_i$ is the pedestrian's position in time $i$, and $S_i$ can be calculated as follows:

$$S_i = M(S_{i-1}, S_{i-2} \ldots S_{i-m}), \quad m \leq i$$

where $M$ is the non-linear mapping from the past trajectories to the current, and $m$ is a constant to measure the length of the observed trajectory. For example, the inference time will increase linearly with sequence length in this way [8, 9]. Suppose the current time is $t$, and $S_{t+3}$ is expected to be predicted. The following calculation process can be performed.

$$\begin{cases} S_{t+1} = T(S_t, S_{t-1} \ldots S_{t-m}) \\ S_{t+2} = T(S_{t+1}, S_t \ldots S_{t-m+1}) \\ S_{t+3} = T(S_{t+2}, S_{t+1} \ldots S_{t-m+2}) \end{cases}$$

Additionally, the surrounding scene features are integrated into the neural network. Based on the problems mentioned above, the proposed model can be formulated as follows:

$$S_{i+n} \cdots S_i = M'(S_{i-1}, S_{i-2} \ldots S_{i-m}, Obstacle)$$

where $Obstacle$ is the surrounding scene features and $M'$ is the non-linear mapping trained in this work.

### 3.2 | Overview

The overall network architecture of Obstacle-Transformer proposed in this paper is illustrated in Figure 2. The inference process can be divided into the following steps.

Step 1: The trajectories are preprocessed, and 'obstacle' is generated by the recent past surrounding trajectories.
Step 2: The features are extracted by patching and embedding according to the processed trajectories and 'obstacle'.
Step 3: Deep features of trajectories and 'obstacle' are extracted by using the Transformer encoder module.
Step 4: The above features are fused by the MLP network to output the predicted trajectory.

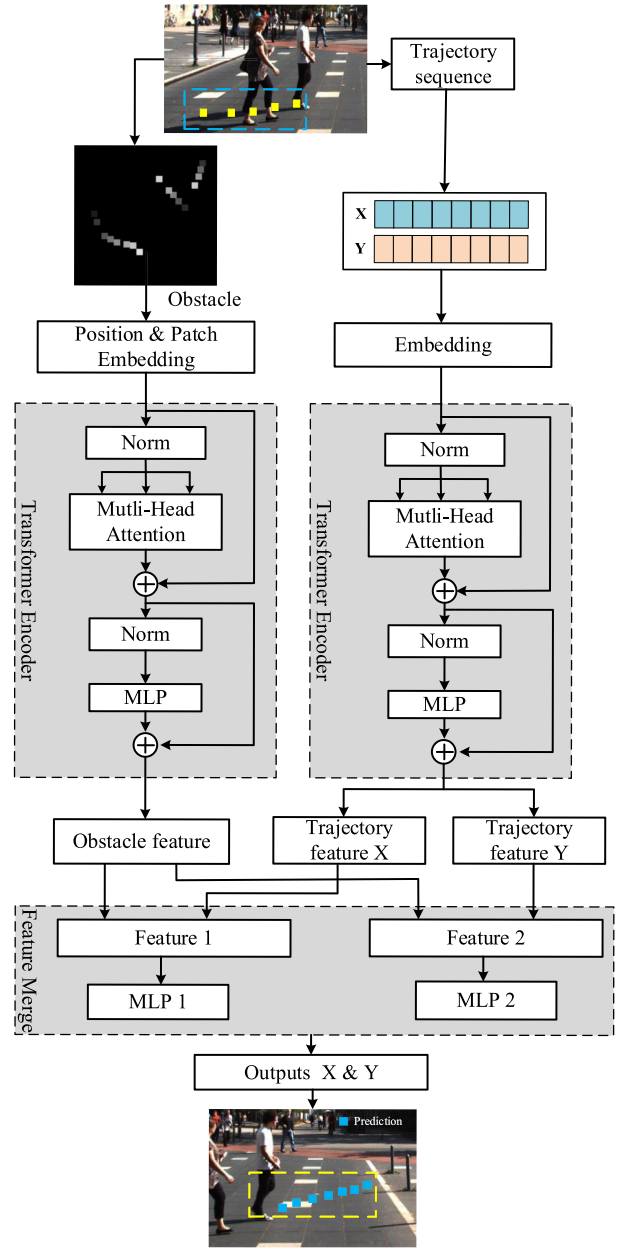

**FIGURE 2** The architecture of the Obstacle-Transformer. The Transformer encoder is utilised in this model to extract features from the trajectory and 'obstacle'. The 'obstacle' is designed by the surrounding trajectories. Compared to the traditional Transformer network, this model integrates the surrounding scene features and has a constant inference time

**TABLE 1** The Pearson correlation coefficients on datasets

| Dataset | Pearson correlation coefficient |
| --- | --- |
| ETH | +0.12 |
| Hotel | −0.17 |
| Univ | −0.10 |

## 3.3 | Trajectory preprocessing

Pedestrian trajectories can be decomposed into two components, $X$ and $Y$. In order to further design the algorithm, the Pearson correlation coefficients of $X$ and $Y$ components are analysed in this work, which can be calculated as the following equation:

$$r = \frac{\sum_{i=1}^{n}(X_i - \overline{X})(Y_i - \overline{Y})}{\sqrt{\text{Var}|X|\text{Var}|Y|}}$$

where $\overline{*}$ is the statistical mean value of the distribution *, and Var $|*|$ is the variance of the distribution *. Pearson correlation coefficients for some of the ETH [29] and UCY [30] datasets are shown in Table 1. The results figure that the $X$ and $Y$ components are uncorrelated; therefore, these two components can be decoupled and predicted separately.

Additionally, three forms of trajectory inputs are proposed. To reduce the difficulty of model learning and improve the convergence speed of model parameters, data differences are constructed as model inputs during training and inference. The methods to generate data differences are shown in Figure 3, method-1 generates a new form of inputs by calculating the position deviation of two adjacent frames, and method-2 by calculating the deviation of each frame with the first frame. Some experiments are conducted in Section 4.3 to evaluate the performance of these methods.

## 3.4 | Obstacle description

To integrate the scene features, the 'obstacle' is proposed by the recent past surrounding trajectories in this paper. The 'obstacle' is a 2-dimensional matrix, which is the result of encoding the positional information of an indeterminate number of trajectories. Three methods are proposed to generate an 'obstacle' in this work, which is shown in Figure 4.

The first method is to encode the last current position of all the surrounding pedestrians. Suppose the $i$th pedestrian's position at time $j$ is $L_{i,j}$, where $L_{i,j} = (x_{i,j}, y_{i,j})$, the current time is $t$, the position of pedestrians at the last current time can be expressed as $L_{i,t-1}$, $i = 0, 1, 2...n$ and $i \neq m$, where $n$ is the number of tracking trajectories, and $m$ is the number of current predicted pedestrians. The 'obstacle' matrix Obs can be calculated in method-1 as follows (Algorithm 1).

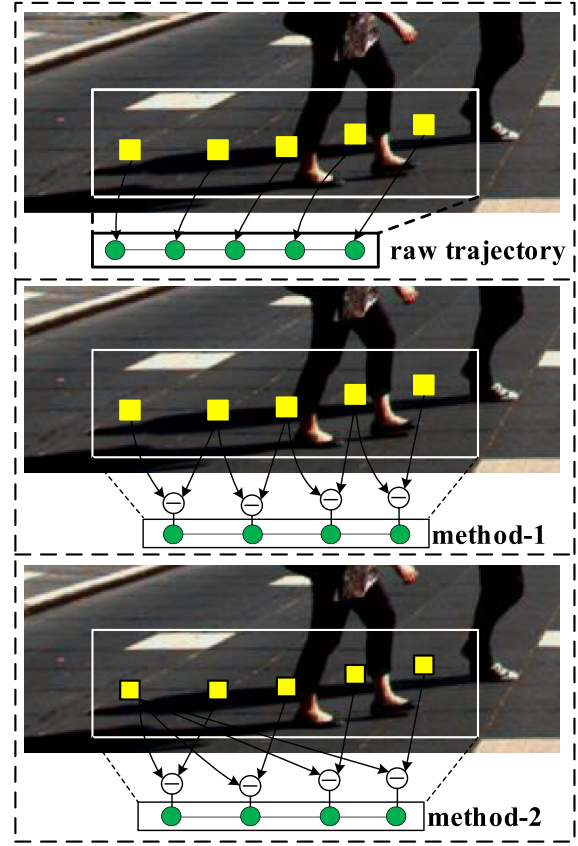

**FIGURE 3** There are three forms of trajectory inputs. The first one uses the raw trajectory as inputs, the method-1 generates a new form of inputs by calculating the position deviation of two adjacent frames, and method-2 by calculating the deviation of each frame to the first frame

**Algorithm 1 Method-1 for generating obstacle.**

**Require:** $L_{i,t-1}$, ($i \in [0, n)$, $i \neq m$), $L_{i,j} = (x_{i,j}, y_{i,j})$
1: Initialise $Obs$ as size $N \times N$ matrix with $Obs[i, j] = 0$, $\forall i, j \in [0, N-1]$
2: **for** $i \in [0, n)$ and $i \neq m$ **do**
3:     **for** $p \in [x_{i,t-1} - 5, x_{i,t-1} + 5]$ **do**
4:         **for** $q \in [y_{i,t-1} - 5, y_{i,t-1} + 5]$ **do**
5:             Update $Obs[p, q] = 1$
6:         **end for**
7:     **end for**
8: **end for**

The second method is to encode $k$ recent past positions of all the surrounding pedestrians. The required positions can be expressed as $L_{i,j}$, $i \in [0, n)$ and $i \neq m$, $j \in [t - k, t - 1]$; the parameters $m$, $n$, $t$ represent the same meaning as above. The algorithm is shown as follows (Algorithm 2).

**Algorithm 2 Method-2 for generating obstacle.**

**Require:** $L_{i,j}$, $i \in [0, n)$ and $i \neq m$, $j \in [t - k, t - 1]$, $L_{i,j} = (x_{i,j}, y_{i,j})$
1: Initialise Obs as size $N \times N$ matrix with

```
Obs[i, j] = 0, ∀i, j ∈ [0, N − 1]
2: for i ∈ [0, n) and i ≠ m do
3:     for j ∈ [t − k, t − 1] do
4:         for p ∈ [x_{i,j} − 5, x_{i,j} + 5] do
5:             for q ∈ [y_{i,j} − 5, y_{i,j} + 5] do
6:                 Update Obs[p, q = 1
7:             end for
8:         end for
9:     end for
10: end for
```

The third method is to encode the $k$ recent past positions of all pedestrians with a linear function. This method has the capability to represent the time interval between the current and the encoded past positions. The 'obstacle' matrix can be calculated as the following algorithm (Algorithm 3).

**Algorithm 3 Method-3 for generating obstacle.**

```
Require: L_{i,j}, i ∈ [0, n) and i ≠ m, j ∈ [t − k,
t − 1], L_{i,j} = (x_{i,j}, y_{i,j})
1: Initialise Obs as size N × N matrix with
Obs[i, j = 0, ∀i, j ∈ [0, N − 1]
2: for i ∈ [0, n) and i ≠ m do
3:     for j ∈ [t − k, t − 1] do
4:         for p ∈ [x_{i,j} − 5, x_{i,j} + 5] do
5:             for q ∈ [y_{i,j} − 5, y_{i,j} + 5] do
6:                 Update Obs[p, q = (j − t + k + 2) /
                   (k + 1)
7:             end for
8:         end for
9:     end for
10: end for
```

### 3.5 | Feature extraction

In this paper, the Transformer encoder and MLP are utilised for feature extraction. Transformer encoder is used to extract deep features from the raw inputs, and MLP is used to merge features and generate predictions.

The capability of Obstacle-Transformer to extract features from sequences mainly lies in the Transformer encoder module. This network is composed of three blocks: multi-head attention modules, feed-forward fully connected modules, and two residual connections. Self-attention can be calculated by the following equation:

$$\text{Attention}(Q, K, V) = \text{softmax}\left(\frac{QK^T}{\sqrt{d_k}}\right) V$$

where $Q$ is the query, $K$ is the key, and $V$ is the value in the self-attention module. $Q$, $K$, and $V$ are calculated by the embedding features $X$. Here are three matrices to learn, $W_Q$, $W_K$, and $W_V$, where $Q = X \cdot W_Q$, $K = X \cdot W_K$, and $V = X \cdot W_V$. Considering the different complexities of the trajectory and 'obstacle', the model is trained with six encoder layers for pursuing trajectory features, four encoder layers for 'obstacle' features, and 8 self-attention heads.

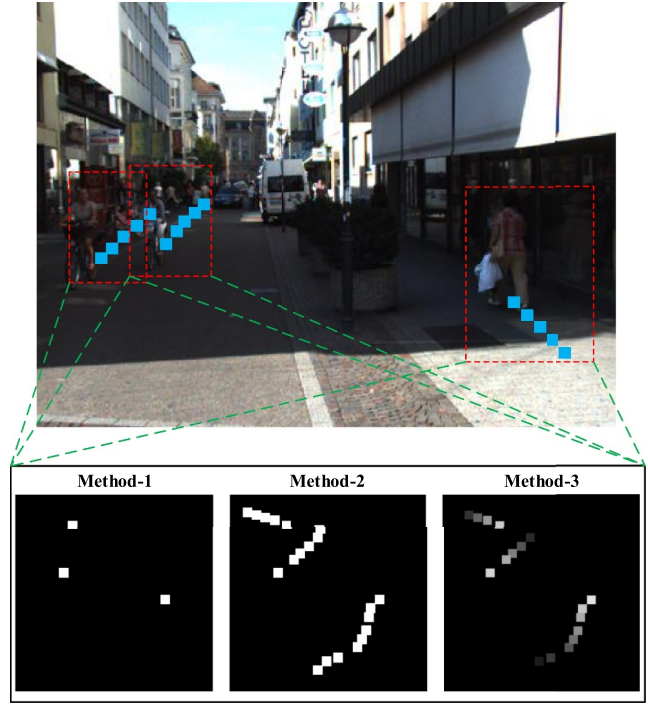

**FIGURE 4** The 'obstacle' is generated by the surrounding trajectories. The following three images show the visualisation results of the three methods. The blue dots are the recent past positions of pedestrians

Compared with Transformer encoder, MLP uses fewer layers for feature fusion. The MLP network proposed in this paper is composed of two fully connected layers (*fc*). The input of the first *fc* is a 768 × 1 feature, and the output is a 256 × 1 feature. The input of the second *fc* is a 256 × 1 feature, and the output is a 12 × 1 feature, which represents the predicted trajectories. The outputs are predicted by a generative approach; in other words, the model generates predictions at multiple times in a single inference, so it can ensure a constant inference time regardless of the length of prediction.

## 4 | EXPERIMENT

To validate the performance of the Obstacle-Transformer model proposed in this paper, experiments are conducted on ETH and UCY trajectory datasets in this section. The dataset and evaluation metrics are introduced in Section 4.1. The comparison of Obstacle-Transformer with other algorithms is shown in Section 4.2. Some ablation experiments are conducted in Section 4.3, and some studies about the influence of prediction length are shown in Section 4.4.

## 4.1 | Dataset and evaluation metrics

ETH is a dataset of pedestrian detection. The testing set contains 1804 images from three video clips. The dataset is captured from a stereo rig mounted on the car with a resolution of 640 x 480, and a framerate of 13–14 frames per second [32]. The UCY dataset consists of real pedestrian trajectories with rich multi-human interaction scenarios captured at 2.5 Hz ($\Delta t = 0.4$ s). It is composed of three sequences (Zara1, Zara2, and UCY), taken in public spaces from the top view [26].

To evaluate the performance of ETH and UCY datasets, Average Displacement Error (*ADE*) and Final Displacement Error (*FDE*) are used as evaluation metrics to measure the prediction error in this work. *Average Displacement Error* can be calculated as follows:

$$ADE = \frac{\sum_{i=1}^{N}\sum_{t=T_{obs+1}}^{T_{pred}}\|L_t^i - \overline{L}_t^i\|_2}{N(T_{pred} - T_{obs})}$$

which is the $L_2$ distance between the true position and the prediction from time $t - T_{obs+1}$ to time $T_{pred}$, and the *FDE* can be calculated as follows:

$$FDE = \frac{\sum_{i=1}^{N}\|L_t^i - \overline{L}_t^i\|_2}{N}$$

which is the $L_2$ distance between the true position and the prediction at the final time $T_{pred}$.

## 4.2 | Quantitative performance

In this subsection, some experiments are conducted to verify the performance of Obstacle-Transformer. Based on the sampling period of 0.4 s for the ETH and UCY datasets, we observed the trajectory for 8 frames (3.2 s) and predicted the next 12 frames (4.8 s). The proposed model in this paper is trained on a single Tesla-P100 GPU, and the experimental results of *ADE* and *FDE* for different algorithms across the datasets are reported in Table 2.

The results show that Obstacle-Transformer achieves competitive performance on the ETH and UCY benchmarks, especially on the metric *ADE*. Meanwhile, there are limitations to the model. Compared with other algorithms in the Hotel dataset, the *FDE* is higher. This may be because when the surrounding trajectories are complex, extracting features directly from the surrounding scenes is not effective enough.

## 4.3 | Ablation experiment

To validate the performance of the 'obstacle' and to assess the differences between the three generating methods of 'obstacle', which are introduced in Section 3.4, some ablation experiments are conducted in this subsection. The frames of observation and prediction are the same as in Section 4.2. The results are shown in Table 3. It shows that the 'obstacle' can integrate scene information to improve the performance of the model. Comparing all three generating methods, method-3 has the best performance in this experiment due to the capability of representing the time interval between the current and past positions.

The effects of the different trajectory input forms introduced in Section 3.3 are studied. The experiments are conducted on the ETH dataset, and the results are reported in Table 4. The results show that taking data differences as inputs can improve the accuracy and convergence speed of the model. Additionally, the performance of method-1 and method-2 is comparable.

**TABLE 2** Quantitative results of different algorithms on ETH and UCY datasets

| Metric | Dataset | Linear | LSTM | S-GAN [22] | S-LSTM [9] | Next [31] | TF [8] | ST-Att. [24] | Obstacle-TF |
|---|---|---|---|---|---|---|---|---|---|
| ADE | ETH | 1.33 | 1.09 | 1.13 | 1.09 | 0.88 | 1.03 | 0.85 | **0.65** |
| | Hotel | 0.39 | 0.86 | 1.01 | 0.79 | 0.36 | 0.36 | 0.32 | **0.27** |
| | Univ | 0.82 | 0.61 | 0.60 | 0.67 | 0.62 | 0.53 | 0.63 | **0.53** |
| | Zara1 | 0.62 | 0.41 | 0.42 | 0.47 | 0.42 | 0.44 | 0.42 | **0.39** |
| | Zara2 | 0.77 | 0.52 | 0.52 | 0.56 | 0.34 | 0.34 | 0.34 | **0.28** |
| | *Average* | 0.79 | 0.70 | 0.74 | 0.72 | 0.52 | 0.54 | 0.51 | **0.42** |
| FDE | ETH | 2.94 | 2.94 | 2.21 | 2.35 | 1.98 | **1.19** | 1.85 | 1.39 |
| | Hotel | 0.72 | 1.19 | 2.18 | 1.76 | 0.76 | 0.71 | **0.66** | 2.08 |
| | Univ | 1.59 | 1.31 | **1.28** | 1.40 | 1.32 | 1.30 | 1.33 | 1.43 |
| | Zara1 | 1.21 | 0.88 | 0.91 | 1.00 | 0.90 | 1.00 | 0.91 | **0.79** |
| | Zara2 | 1.48 | 1.11 | 1.11 | 1.17 | 0.75 | 0.76 | 0.73 | **0.67** |
| | *Average* | 1.59 | 1.52 | 1.54 | 1.54 | 1.14 | 1.17 | **1.10** | 1.27 |

*Note*: The bold values is the best performance in each row, which mean the corresponding algorithm has the best performance in this dataset.
Abbreviations: ADE, Average Displacement Error; FDE, Final Displacement Error; LSTM, Long Short-Term Memory; TF, Transformer.

**TABLE 3** Ablation experiments for 'obstacle'

| Metric | Method | ETH | Hotel | Univ | Zara1 | Zara2 | Average |
|---|---|---|---|---|---|---|---|
| ADE | No obstacle | 1.08 | 0.41 | 0.57 | 0.42 | 0.39 | 0.59 |
| | Method-1 | 0.82 | 0.30 | 0.54 | 0.45 | 0.35 | 0.49 |
| | Method-2 | 0.78 | 0.31 | 0.54 | 0.43 | 0.30 | 0.47 |
| | Method-3 | **0.65** | **0.27** | **0.53** | **0.39** | **0.28** | **0.42** |
| FDE | No obstacle | 1.67 | 2.25 | 1.68 | 0.92 | 0.79 | 1.46 |
| | Method-1 | 1.49 | 2.09 | 1.58 | 0.82 | 0.77 | 1.35 |
| | Method-2 | 1.47 | 2.03 | 1.49 | 0.87 | 0.71 | 1.31 |
| | Method-3 | **1.38** | **2.08** | **1.43** | **0.79** | **0.67** | **1.27** |

*Note*: The bold values is the best performance in each row, which mean the corresponding algorithm has the best performance in this dataset.

**TABLE 4** Ablation experiments for the inputs

| Method | ADE | FDE |
|---|---|---|
| Raw trajectory | 1.34 | 3.79 |
| Method-1 | **0.65** | 1.38 |
| Method-2 | 0.71 | **1.28** |

*Note*: The bold values is the best performance in each row, which mean the corresponding algorithm has the best performance in this dataset.

**TABLE 5** The influence of prediction length for inference time (s)

| Length | Transformer | Obstacle-TF |
|---|---|---|
| 12 | 0.684 | 0.169 |
| 16 | 0.920 | 0.153 |
| 20 | 1.181 | 0.167 |
| 24 | 1.545 | 0.164 |
| 28 | 1.873 | 0.159 |
| 32 | 2.113 | 0.167 |

Abbreviation: TF, Transformer.

## 4.4 | Length of prediction

In this subsection, the influences of prediction length on inference time and accuracy are analysed. The observation sequences varied from 16 frames (4.8 s) to 32 frames (12.8 s) with a step of 1.8 s. The experiments were conducted on the ETH datasets.

The experiment compares our Obstacle-Transformer with the raw Transformer, and the influence of observation length on the inference time is shown in Table 5. The results figure that the proposed model in this paper has a constant inference time when the prediction length increases. Additionally, compared with Transformer, our method has a better real-time performance.

Additionally, the accuracy under different prediction lengths of the Transformer, LSTM, and Obstacle-Transformer is reported in Table 6. The results figure that Obstacle-Transformer has better accuracy when the prediction length

**TABLE 6** The influence of prediction length for accuracy

| Length | Method | ADE | FDE |
|---|---|---|---|
| 16 | LSTM | 1.15 | 2.72 |
| | Transformer | 0.95 | 2.15 |
| | Obstacle-TF | **0.79** | **2.01** |
| 20 | LSTM | 1.64 | 3.99 |
| | Transformer | 1.27 | **2.90** |
| | Obstacle-TF | **1.18** | 2.91 |
| 24 | LSTM | 2.29 | 5.55 |
| | Transformer | 1.66 | 3.76 |
| | Obstacle-TF | **1.58** | **3.48** |
| 28 | LSTM | 3.07 | 7.46 |
| | Transformer | **2.27** | 5.09 |
| | Obstacle-TF | 3.06 | **5.01** |
| 32 | LSTM | 4.13 | 9.96 |
| | Transformer | **2.98** | **4.52** |
| | Obstacle-TF | 3.38 | 5.15 |

*Note*: The bold values is the best performance in each row, which mean the corresponding algorithm has the best performance in this dataset.
Abbreviations: LSTM, Long Short-Term Memory; TF, Transformer.

is short. However, the accuracy of Transformer is better than that of the others, while the prediction length becomes longer. This is because the algorithm predicts the deviation of two adjacent frames, and when the predicted trajectory becomes longer, the cumulative error will increase.

## 5 | CONCLUSIONS

An Obstacle-Transformer model has been proposed for predicting trajectories in this work. By designing the 'obstacle' in this model, the surrounding scene features have been integrated into the network. This method has made progress in trajectory prediction of some interactive scenes. Moreover, the 'obstacle' was generated by the surrounding trajectory rather than images or point cloud scenes, which makes the method more applicable to a wider range of scenarios. Additionally, an MLP feature extraction method has been proposed in this work, which optimises the inference time to a constant and is independent of the prediction length. Finally, experiments were conducted to verify the real-time performance and accuracy of the Obstacle-Transformer model. Also, the ablation experiments show that the architecture of the 'obstacle' might influence the feature extraction and trajectory prediction. Our further work will focus on this problem and study its inner principle.

**ACKNOWLEDGEMENT**
This work was supported in part by the National Science Foundation of China under Grant No. 62033005, Grant No. 62022030, Grant No. 62106062, in part by the Natural Science

Foundation of Heilongjiang Province under Grant No. ZD2021F001, in part by the Fundamental Research Funds for the Central Universities (Grant No. HIT.OCEF. 2021005), and in part by the China Postdoctoral Science Foundation under Grant No. 2021M701007 and Grant No. 2021TQ0091.

## CONFLICT OF INTEREST

The authors declare that there is no conflict of interest that could be perceived as prejudicing the impartiality of the research reported.

## DATA AVAILABILITY STATEMENT

The data that support the findings of this study are available from the corresponding author upon reasonable request.

## ORCID

*Chengwei Wu*  https://orcid.org/0000-0001-9600-0205